\documentclass[runningheads, envcountsame, a4paper]{llncs}
\usepackage[misc]{ifsym}
\usepackage{graphicx}
\usepackage[colorinlistoftodos]{todonotes}
\usepackage[hyphens]{url}
\usepackage{hyperref}

\usepackage{listings}

\usepackage{color}
\definecolor{deepblue}{rgb}{0,0,0.5}
\definecolor{deepred}{rgb}{0.6,0,0}
\definecolor{deepgreen}{rgb}{0,0.5,0}

\begin{document}

\title{GAMA: a General Automated Machine learning Assistant}

\author{Pieter Gijsbers[\Letter] \and
Joaquin Vanschoren}
\tocauthor{Pieter~Gijsbers and Joaquin~Vanschoren}
\toctitle{GAMA: a General Automated Machine learning Assistant}
\authorrunning{P. Gijsbers and J. Vanschoren}

\institute{Eindhoven University of Technology, Eindhoven, Netherlands\\
\email{p.gijsbers@tue.nl}
}
\maketitle              
\begin{abstract}
The General Automated Machine learning Assistant (GAMA) is a modular AutoML system developed to empower users to track and control how AutoML algorithms search for optimal machine learning pipelines, and facilitate AutoML research itself. In contrast to current, often black-box systems, GAMA allows users to plug in different AutoML and post-processing techniques, logs and visualizes the search process, and supports easy benchmarking. It currently features three AutoML search algorithms, two model post-processing steps, and is designed to allow for more components to be added.

\end{abstract}

\section{Introduction}
Automated Machine Learning (AutoML) aims to automate the process of building machine learning models, for instance, by automating the selection and tuning of preprocessing and learning algorithms in machine learning pipelines. 
In recent years, many AutoML systems have been developed, such as Auto-WEKA~\cite{autoweka}, auto-sklearn~\cite{autosklearn}, TPOT~\cite{tpot} and ML-Plan~\cite{mlplan}.
They vary in the types of pipelines they build (e.g. fixed or variable length), how they optimize them (e.g. using evolutionary or Bayesian optimization), and whether or how they employ meta-learning (e.g. warm-starting) or post-processing (e.g. ensembling).

We demonstrate\footnote{A video demonstration can be found at \url{https://youtu.be/angsGMvEd1w}} a new open-source AutoML system,  GAMA\footnote{Code and documentation can be found at \url{https://github.com/PGijsbers/gama/}}~\cite{gama},
which distinguishes itself by it modularity (allowing users to compose AutoML systems from sub-components), extensibility (allowing new components to be added), transparency (tracking and visualizing the search process to better understand what the AutoML system is doing), and support for research, such as integration with the AutoML benchmark~\cite{amlb}. The main difference to our earlier publication (\cite{gama}) is the redesign to allow for a modular AutoML pipeline and the addition of a graphical user interface.

As such, it caters to a wide range of users, from people without a deep machine learning background who want an easy-to-use AutoML tool, to those who want better control and understanding of the AutoML process, and especially researchers who want to perform systematic AutoML research.

Currently, three different search algorithms and two post-processing techniques are available, but we welcome and plan to include more techniques in the future. For novice users, GAMA offers a default configuration shown to perform well in our benchmarks.

\section{System Overview}

\subsection*{Modular AutoML Pipeline} 
Rather than prescribing a specific combination of AutoML techniques, GAMA allows users to combine different search and post-processing algorithms into a flexible AutoML `pipeline' that can be tuned to the problem at hand. 

There are three optimization algorithms currently implemented in GAMA to search for optimal machine learning pipelines: random search~\cite{randomsearch}, an asynchronous successive halving algorithm (ASHA)~\cite{asha} which uses low-fidelity estimates to filter out bad pipelines early, and an asynchronous multi-objective evolutionary algorithm.

After the pipeline search has completed, a post-processing technique will be executed to construct the final model. It is currently possible to either train the single best pipeline or create an ensemble out of pipelines evaluated during search, as described in~\cite{rich}.
In subsequent work, we plan to expand the number of search and post-processing techniques available out-of-the-box.

Listing~\ref{code:automlpipeline} shows how to configure GAMA with non-default search and postprocessing methods and use it as a drop-in replacement for scikit-learn estimators.\footnote{An always up-to-date version of this listing can be found at \url{https://pgijsbers.github.io/gama/master/citing.html}} 

New AutoML algorithms or variations to existing ones can be included and tested with relative ease. 
For instance, each of the search algorithms described above has been implemented and integrated in GAMA with less than 170 lines of code, and they can all make use of shared functions for logging, parallel pipeline evaluation and adhering to runtime constraints.
It also allows users to research other questions, such as how to choose the search algorithm for AutoML.

\begin{lstlisting}[language=Python, float=tp,
  floatplacement=tbp, abovecaptionskip=-5pt,
  caption={Configuring an AutoML pipeline with GAMA} \label{code:automlpipeline}]
from gama import GamaClassifier
from gama.search_methods import AsynchronousSuccessiveHalving
from gama.postprocessing import EnsemblePostProcessing

automl = GamaClassifier(
    search=AsynchronousSuccessiveHalving(),
    post_processing=EnsemblePostProcessing()
)
automl.fit(X, y)
automl.predict(X_test)
automl.fit(X_test, y_test)

\end{lstlisting}

\subsection*{Interface}
GAMA comes with a graphical web interface which allows novice users to start and configure GAMA.
Moreover, it visualizes the AutoML process to enable researchers to easily monitor and analyse the behavior of specific AutoML configurations.

\begin{figure}[t]
  \includegraphics[width=\linewidth]{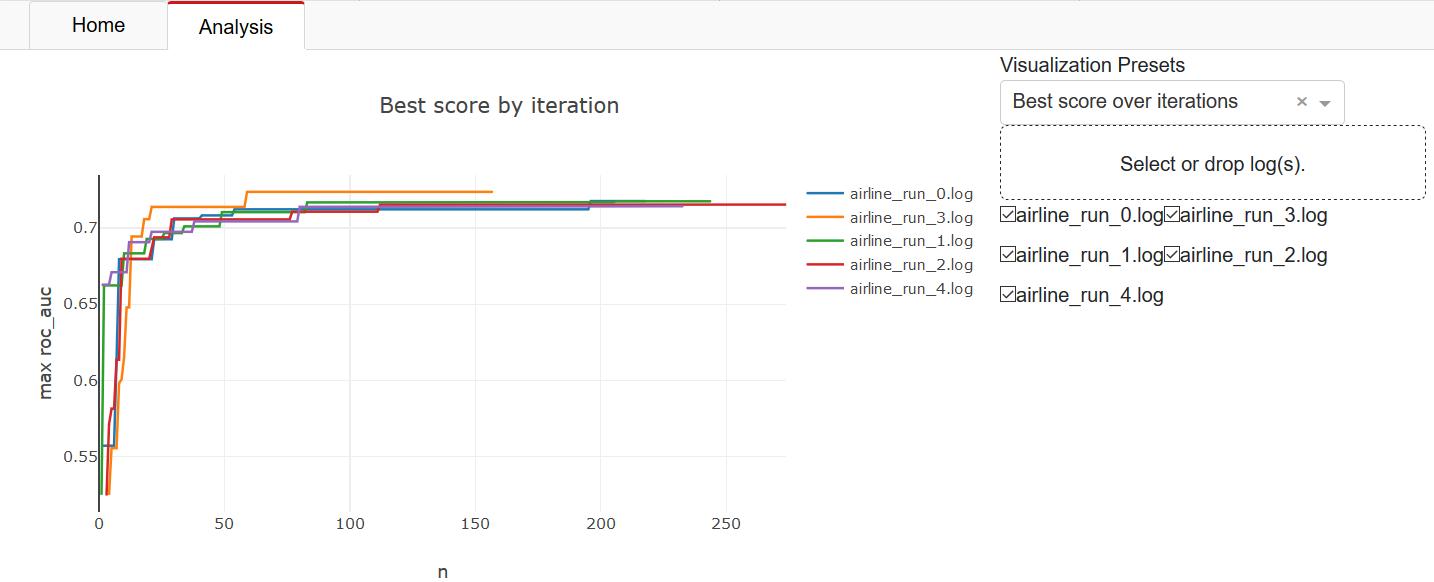}
  \caption{Visualization of logs}
  \label{fig:logvis}
\end{figure}

GAMA logs the creation and evaluation of each pipeline, including meta-data such as creation time and evaluation duration. For pipelines created through evolution, it also records the parent pipelines and how they differ. One can also compare multiple logs at once, creating figures such as Figure~\ref{fig:logvis} that shows the convergence rate of five different GAMA runs over time on the airline dataset\footnote{\url{https://www.openml.org/d/1169}}.

\subsection*{Benchmarking}

\begin{figure}[t]
  \includegraphics[width=\linewidth]{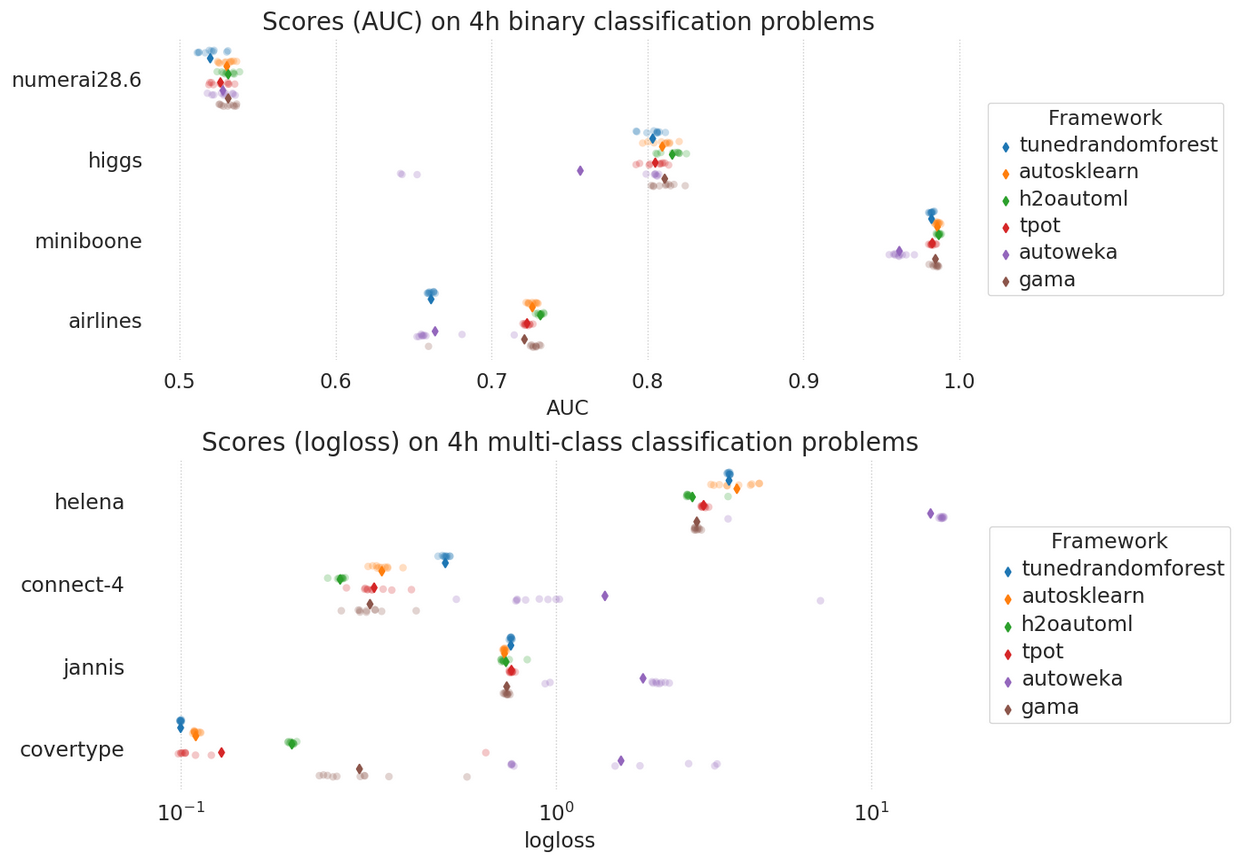}
  \caption{Performance benchmark results}
  \label{fig:benchbin}
\end{figure}

GAMA in integrated with the open-source AutoML Benchmark introduced in~\cite{amlb}. Figure~\ref{fig:benchbin} shows the results of running GAMA with its default settings some of the biggest and most challenging datasets for which each other framework had results in the original work.\footnote{Although we could not run these experiments on the same (AWS) hardware, we took care to use the same computational constraints.}
The full and latest results will be made available in the GAMA documentation.

\section{Related work}
GAMA compares most closely to auto-sklearn and TPOT as they also optimize scikit-learn pipelines. Auto-sklearn and GAMA both implement the same ensembling technique~\cite{rich}.
GAMA and TPOT both feature evolutionary search with NSGA2 selection~\cite{nsga2}, although GAMA's implementation uses asynchronous evolution, which is often faster.
While TPOT and auto-sklearn have a fixed AutoML pipeline, they do allow modifications to their search space.
To the best of our knowledge, GAMA is the only AutoML framework 
that offers a modular and extensible composition of AutoML systems, and extensive support for AutoML research.

\section{Conclusion}
In this proposal we presented GAMA, an open-source AutoML tool which facilitates AutoML research and skillful use through its modular design and built-in logging and visualization.
Novice users can make use of the graphical interface to start GAMA, or simply use the default configuration which is shown to generate models of similar performance to other AutoML frameworks.
Researchers can leverage GAMA's modularity to integrate and test new AutoML search procedures in combination with other readily available building blocks, and then log, visualize, and analyze their behavior, or run extensive benchmarks. In future work, we aim to integrate additional search techniques as well as extend the AutoML pipeline with additional steps, such as warm-starting the pipeline search with meta-data.

\section*{Acknowledgements}
This software was developed with support from the Data Driven Discovery of Models (D3M) program run by DARPA and the Air Force Research Laboratory.

\end{document}